\documentclass[runningheads]{llncs}
\usepackage{graphicx}
\graphicspath{ {images/} }

\usepackage{tikz, pgfplots}
\usepackage{comment} 
\usepackage{booktabs}
\usepackage{tabularx}
\usepackage[utf8]{inputenc}
\usepackage{amsmath,amssymb} 
\newcommand{\norm}[1]{\left\lVert#1\right\rVert}
\DeclareMathOperator*{\argmin}{argmin} 
\newcommand*\samethanks[1][\value{footnote}]{\footnotemark[#1]}
\usepackage{color}

\usepackage[width=122mm,left=12mm,paperwidth=146mm,height=193mm,top=12mm,paperheight=217mm]{geometry}

\begin{document}
\pagestyle{headings}
\mainmatter
\def\ECCVSubNumber{5530}  

\title{Semantic Relation Preserving Knowledge Distillation for Image-to-Image Translation} 

\titlerunning{Semantic Relation Preserving KD}
%
\author{Zeqi Li\inst{}\thanks{Equal contribution}, 
Ruowei Jiang\inst{}\samethanks, \and
Parham Aarabi\inst{}}
\authorrunning{Li Z., Jiang R., et al.}
%
\institute{ModiFace\\
\email{\{lizeqi,irene,parham\}@modiface.com}}
\maketitle

\begin{abstract}
Generative adversarial networks (GANs) have shown significant potential in modeling high dimensional distributions of image data, especially on image-to-image translation tasks. However, due to the complexity of these tasks, state-of-the-art models often contain a tremendous amount of parameters, which results in large model size and long inference time. In this work, we propose a novel method to address this problem by applying knowledge distillation together with distillation of a semantic relation preserving matrix. This matrix, derived from the teacher's feature encoding, helps the student model learn better semantic relations. In contrast to existing compression methods designed for classification tasks, our proposed method adapts well to the image-to-image translation task on GANs. Experiments conducted on 5 different datasets and 3 different pairs of teacher and student models provide strong evidence that our methods achieve impressive results both qualitatively and quantitatively.
\keywords{Knowledge Distillation, Generative Adversarial Networks, Image-to-Image Translation, Model Compression}
\end{abstract}

\section{Introduction}

Generative adversarial networks (GANs) \cite{goodfellow2014generative} have presented significant potential in modeling high dimensional distributions of image data, on a variety of visual tasks. Many of these tasks, such as style-transfer \cite{zhu2017unpaired,isola2017image} and super-resolution \cite{ledig2017photo}, are considered to be image-to-image translation tasks, in which we train a model to map images from one domain to another. The community has shown success in researching solutions to generate high fidelity images \cite{brock2018large,shaham2019singan} and dealing with unpaired data \cite{zhu2017unpaired}. The success in these works has also led to a popular trend of developing mobile applications based on generative models. However,  little work has been done in making these models efficient on mobile devices.  As a result, the state-of-the-art GAN models are often large and slow on resource-limited edge devices. For instance, a CycleGAN \cite{zhu2017unpaired} model needs 2.69 seconds to process one image of resolution 256x256 on a single CPU core of Intel(R) Xeon(R) E5-2686, with the model being 44M large.

With achievements of convolutional neural networks (CNNs), many works \cite{han2015deep,sandler2018mobilenetv2,howard2017mobilenets,he2018amc,howard2019searching} for model compression have been proposed to improve model efficiency in a variety of computer vision tasks including classification, object detection and semantic segmentation. In 2016, Han et al. \cite{han2015deep} proposed a three-stage pipeline that first prunes the model by cutting down less important connections and then quantizes the weights and applies Huffman encoding. They successfully reduced AlexNet \cite{krizhevsky2012imagenet} and VGG-16 \cite{simonyan2014very} by 35x to 49x on the ImageNet dataset \cite{deng2009imagenet}. This method, with a complex training pipeline, requires a great amount of manual efforts in each stage.  In \cite{sandler2018mobilenetv2,howard2017mobilenets}, efforts have been dedicated to improving model efficiency by redesigning convolutional layers into separable convolutional layers. Redesigning network architecture often requires domain experts to explore the large design space and conduct a significant amount of experiments. Later works such as \cite{he2018amc,howard2019searching}, have leveraged techniques in neural architectural search and reinforcement learning to efficiently reduce the amount of such manual efforts by performing pruning and network designing based on a trained agent's predictions. Upon successful results in compressing networks for classification tasks,  research works \cite{chen2019detnas,redmon2018yolov3,Liu_2019_CVPR} have further extended the aforementioned techniques to object detection and semantic segmentation. 

\begin{figure}[t]
\label{demonstration}
\includegraphics[width=0.95\textwidth]{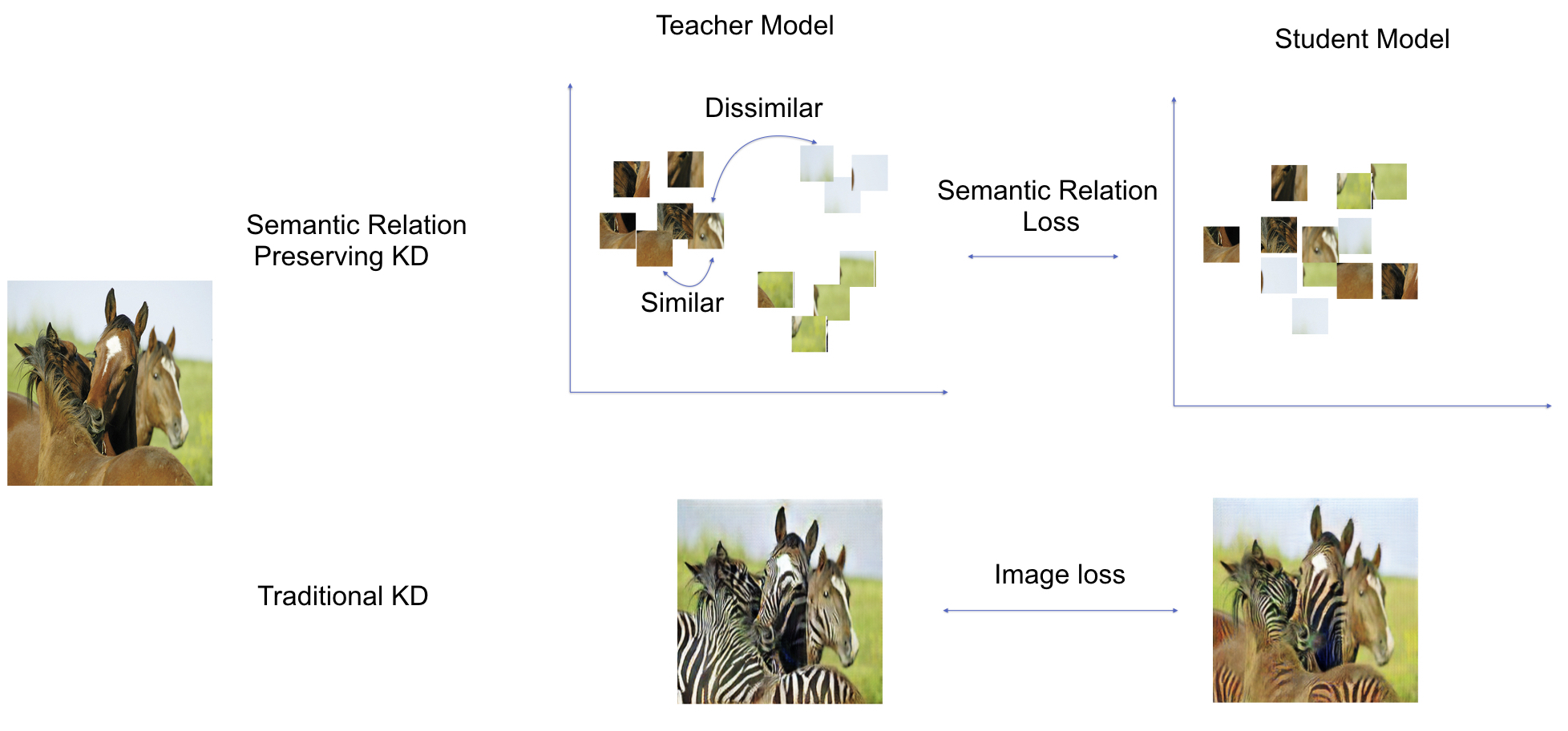}
\caption{A visualization of our proposed idea. In the top row, we show that our proposed method transfers the semantic relations learned in a teacher model to a student model. In high dimensional space, feature encoding for pixels of the same semantic class may locate closer. The bottom row shows how traditional knowledge distillation would work on image-to-image translation tasks}
\end{figure}

However, the aforementioned solutions do not adapt well to GANs, as GANs typically demand excessive amounts of training processes and manual design efforts. The training of generative adversarial networks is usually harder and less stable due to the design of alternating training strategy for the discriminator and the generator. Therefore, we explore methods that not only improve the model's efficiency but also provide guidance while training. Hinton et al. \cite{hinton2015distilling} reinvented the concept of knowledge distillation to transfer the dark knowledge from an ensemble teacher model to a single student model, which demonstrated the potential of utilizing knowledge distillation in model compression. In this setting, inexplicit and intermediate information such as probability distribution from the teacher's network can be leveraged at training time to guide the student. Given the intuition of this concept, knowledge distillation naturally fits our objective of compressing a GAN generator with a guided training procedure.

In this work, we apply knowledge distillation on image-to-image translation tasks and further propose a novel approach to distill information of semantic relationships from teacher to student. Our hypothesis is that, given a feature tensor, feature pixels of the same semantic class may have similar activation patterns while feature pixels of different semantic classes may be dissimilar. To better illustrate our idea, we provide a visualization in \textbf{Fig. \ref{demonstration}}.  For example, on the horse-to-zebra task, feature tensors of horses may locate closer but far from other background pixels such as sky and grass in high dimensional space. A well-trained teacher model is able to capture these correlations better among different semantic pixels at both dataset and image level. We will also demonstrate evidence to support this intuition in Methods.

Our main contributions of this work are:
\begin{itemize}
  \item[    $\bullet$]We present a novel method of applying knowledge distillation in compressing GAN generators on image-to-image translation task by distilling the semantic relations. Th student model's pixel pairwise similarities are trained in a supervised setting by the teacher's. 
\item[       $\bullet$] We experimentally demonstrate the potential of this method on 5 different image-to-image translation benchmark datasets. Our results, both qualitatively and quantitatively, evidently show that our method trains the student model to be on par with and sometimes better than the original teacher model.
\end{itemize}
\section{Related Work}
\subsection{GANs for Image-to-Image Translation}
Along with the success of GANs in modeling high dimensional data, image-to-image translation tasks are dominated by GANs nowadays due to GANs' superiority in generating images of high fidelity and extendibility on different data domains. In \cite{isola2017image}, authors proposed a model known as Pix2Pix applying conditional GANs on paired image-to-image translation tasks such as transferring from sketches/semantic labels to photos. A subsequent work CycleGAN \cite{zhu2017unpaired}, tackling unpaired image-to-image translation tasks between two domains, proposed to construct two generators transferring images in both directions and enforce an additional cycle consistency loss during the training. StarGAN \cite{choi2018stargan} has further extended the capability of CycleGAN to the multi-domain translation by adding a domain-specific attribute vector in the input while training the generators. 

\subsection{Semantic Relation Preserving Knowledge Distillation}
There has been a long line of efforts dedicated to transferring knowledge from a teacher model to a student model. Hinton et al. \cite{hinton2015distilling} reinvented the concept of knowledge distillation in which a single student model learns the knowledge from an ensemble of separately trained models. Comparing to one-hot output, the information contained within a teacher's soft logits provides more concrete knowledge and helps guide the training of a student model. In addition to classification tasks, this idea has also been widely adopted in numerous computer vision tasks such as object detection and semantic segmentation \cite{NIPS2017_6676,liu2019structured}.

Recently, it is observed that learning class relationship enhances model performance non-trivially in various problems. Many works \cite{chen2018darkrank,peng2019correlation,park2019relational,tung2019similarity} have shown progress in applying similarity and relational learning in a knowledge distillation setting.  In \cite{park2019relational} and \cite{peng2019correlation}, they both demonstrated that correlation among instances can be transferred and well learned in a student model through geometric similarity learning of multiple instances.  In \cite{tung2019similarity}, they demonstrated empirically that similar activation patterns would appear on images of the same class (e.g. dog). Based on this observation, they proposed to guide the student with a similarity matrix of image instances calculated as the outer product of the teacher's feature encoding of certain layers. However, on the image-to-image translation tasks, image-wise relationships do not give comprehensive information as they are typically images from the same class (e.g. horses, zebras). Might similar correlation patterns exist among semantic pixels? In this work, we explore the idea to retain pixel-wise semantic relation in the student model, by transferring this knowledge from the teacher.
\subsection{Model Compression on GANs}
Image-to-image translation tasks using generative models are essentially different from classification tasks with discriminative models. Traditional model compression approaches are designed for classification tasks, which do not adapt well to GANs trivially. Work \cite{aguinaldo2019compressing} applied KD to compress the GAN generator by enforcing a joint loss of pixel-wise loss and adversarial loss with a shared discriminator, with a focus on the unconditional image generation task. Another work \cite{shu2019co} devoted effort to compressing GAN models through a co-evolutionary strategy of the two generators in CycleGAN \cite{zhu2017unpaired}, resulting in a method that efficiently eliminates redundant convolutional filters. However, it requires external effort to maintain the quality of generated images by controlling the model compression ratio and other hyper-parameters. 

In this work, we aim to tackle the GAN compression problem on the image-to-image translation task. Our proposed KD method inspired by [29] from the classification task significantly reduces the amount of effort needed for hyper-parameter tuning and achieves better image fidelity while realizing effective compression by leveraging the semantic relation similarity between feature pixels of images from a well-trained teacher model. 

\begin{figure}[t!]
\centering
\includegraphics[width=\textwidth]{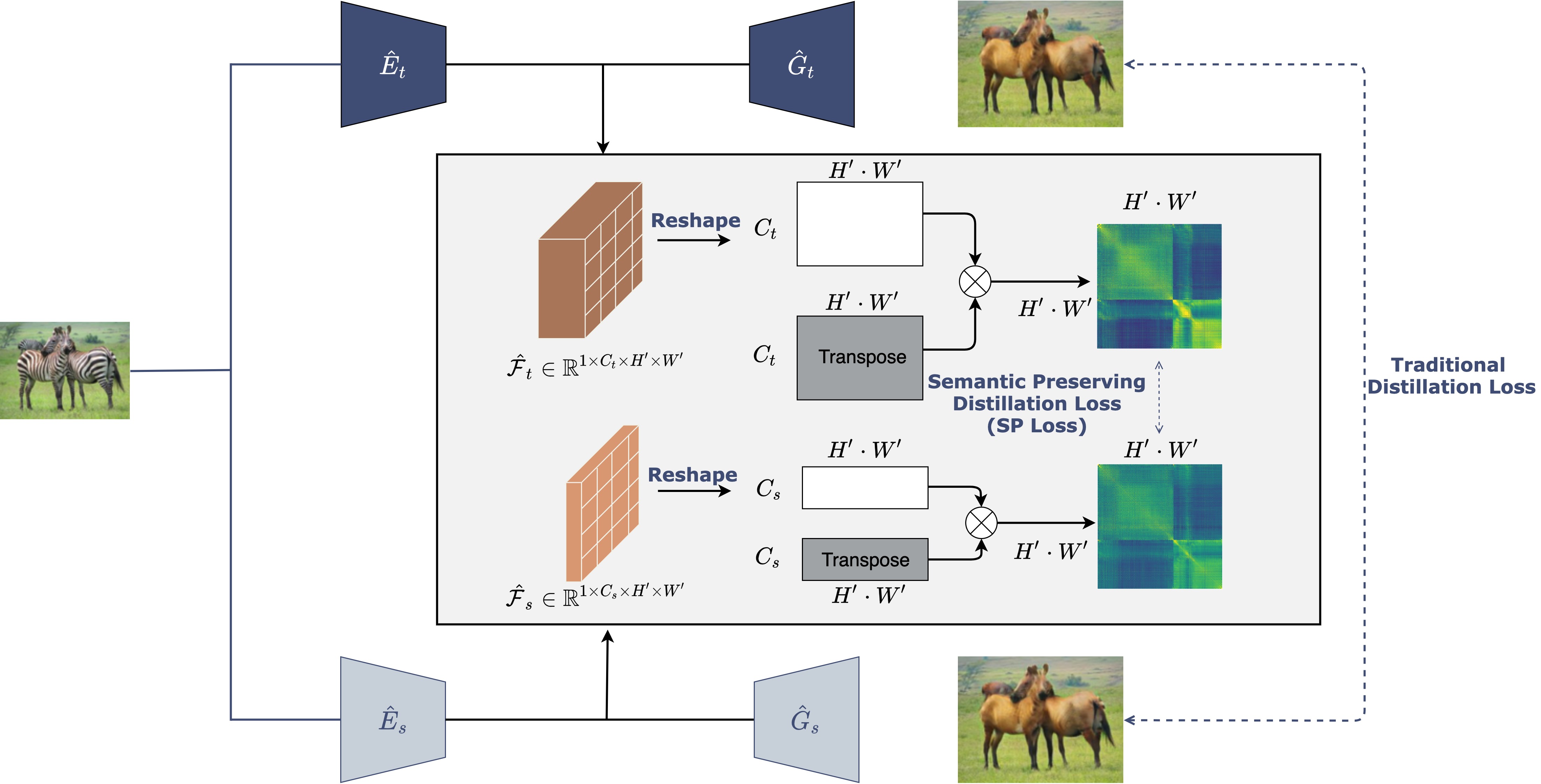}
\caption{An overview of our training pipeline. The semantic relation activation matrix is calculated as the outer product of the feature encoding. A distillation loss is used to compare the teacher's activation matrix and the student's}
\label{training_net}
\end{figure}

\section{Methods}
The goal of this work is to improve GANs efficiency by utilizing knowledge distillation in compressing the generator. As discussed in the Introduction, the training of GANs is challenging. In addition to the vanilla knowledge distillation loss, we separate each generator into one encoder and one decoder and formulate a semantic preserving loss based on the feature encoding produced by the encoder. In \textbf{Fig. \ref{training_net}}, we present an overview of our distillation strategies in preserving semantic relationships. At an intermediate layer, we represent semantic relations by calculating pairwise activation similarities on pixels of the feature encoding and transfer the knowledge via a distillation loss on the similarity matrices. This loss can be added in addition to traditional distillation loss on the final generated images. In this section, we will discuss details about how we apply vanilla knowledge distillation and semantic preserving distillation on GANs.

\subsection{Vanilla Knowledge Distillation on GANs}
In traditional knowledge distillation, the task is formulated as:

\begin{equation}
\theta_s = \argmin_{\theta}  {\mathbb{E}_{x_i,y_i}} \Big[(1-\alpha) \mathcal{L}(y_i, f_{\theta}(x_i)) + \alpha \mathcal{L}(f_t(x_i), f_{\theta}(x_i)) \Big],
\label{kd_eq}
\end{equation}

where $y_i$ denotes the ground truth for input $x_i$,  $f_\theta(x_i)$ and $f_t(x_i)$ denote the student model output and teacher model output respectively. $n$ is the number of inputs and $\alpha$ is a hyper-parameter to balance between teacher's output and the ground truth. 
\textbf{Equation (\ref{kd_eq})} encourages the network to minimize two terms: 1) the loss between ground truth and student's output, and 2) the loss between the teacher's output and the student's output.  The second part of the objective function is designed to help the student learn inexplicit knowledge on different tasks. For example, on a classification task, soft logits with temperature control are matched between the student and the teacher to encourage the student to mimic the teacher.

In the setting of generative adversarial training, an example approach to applying knowledge distillation would be introducing another minimax game between the teacher's generated images ($G_t(x)$) and the student's ($G_s(x)$):
\begin{multline}
\label{gan_kd}
\min_{G_s}  \max_{\mathcal{D}_s} V(G_s, \mathcal{D}_s) = \alpha \Big( {\mathbb{E}}_{\scriptscriptstyle{y\sim \mathcal{P}_{data}(y)}}[\log \mathcal{D}_s(y)] +  {\mathbb{E}}_{\scriptscriptstyle{x \sim \mathcal{P}_{data}(x)}}[\log (1 - \mathcal{D}_s(G_s(x))] \Big) \\
 +(1-\alpha)\mathcal{L}_{KD},
\end{multline}
where
\begin{equation}
\mathcal{L}_{KD} = {\mathbb{E}}_{ \scriptscriptstyle{y \sim \mathcal{P}_{data}(G_t(x))}}[\log \mathcal{D}_s'(y)] +  {\mathbb{E}}_{\scriptscriptstyle{x \sim \mathcal{P}_{data}(x)}}[\log (1 - \mathcal{D}_s'(G_s(x))],
\end{equation} 
Subscript $t$ and $s$ indicate components of the teacher and the student. $\mathcal{D}_s$ is the discriminator for the student's output and real images while $\mathcal{D}_s'$ differentiate student's output and teacher's. $x$ and $y$ are real images from its respective class. 

In our preliminary experiments, we tried using such adversarial loss between teacher and student's output, but we found this strategy is unstable and difficult to train. Besides, we did not observe improved performance on converged experiments. Previous works \cite{zhu2017unpaired,isola2017image} have shown the benefits of mixing GAN objective with other traditional losses such as L1. Therefore, we apply vanilla knowledge distillation by computing a traditional reconstruction loss comparing teacher's and student's output. For example in CycleGAN \cite{zhu2017unpaired}, the original loss is weighted among two GAN losses and one cycle consistency loss. We add the distillation loss only on cycle consistency loss which is an L1 norm loss. Our vanilla knowledge distillation setting has the following objective:
\begin{multline}
\label{cyclegan_loss}
\mathcal{L}(G_s, F_s, D_X, D_Y) = \mathcal{L}_{adv}(G_s, D_Y, X, Y)+\mathcal{L}_{adv}(F_s, D_X,  Y, X) \\
+ \lambda(\alpha \mathcal{L}_{cyc}(G_s, F_s, X, Y) + (1 -  \alpha) \mathcal{L}_{cyc}(G_s, F_s, X_t, Y_t)),
\end{multline}
where $\mathcal{L}_{adv}$ is the adversarial loss and $\mathcal{L}_{cyc}$ is the reconstruction loss. Also, $G_s$ and $F_s$ denote generators transferring from style class $X$ to $Y$ and $Y$ to $X$ respectively. Accordingly, $X_t$ and $Y_t$ are teacher generated reconstruction images. $\lambda$ is the balancing coefficient. Notations are adapted from \cite{zhu2017unpaired}. We also apply similar settings in Pix2Pix \cite{isola2017image} training. The detailed objective function is described in Supplementary.

\subsection{Semantic Preserving Loss}
\textbf{Notation.} We consider a generator G to be composed by two parts: an encoder $\hat E$ that encodes the input images and a generator $\hat G$ that decodes and generates the output images. We note $y_i$ to be the output image of \textit{i-th} input $x_i$ where $y_i = G(x) = \hat G ( \hat E(x_i)$.

\begin{figure}[t]
\includegraphics[width=\textwidth]{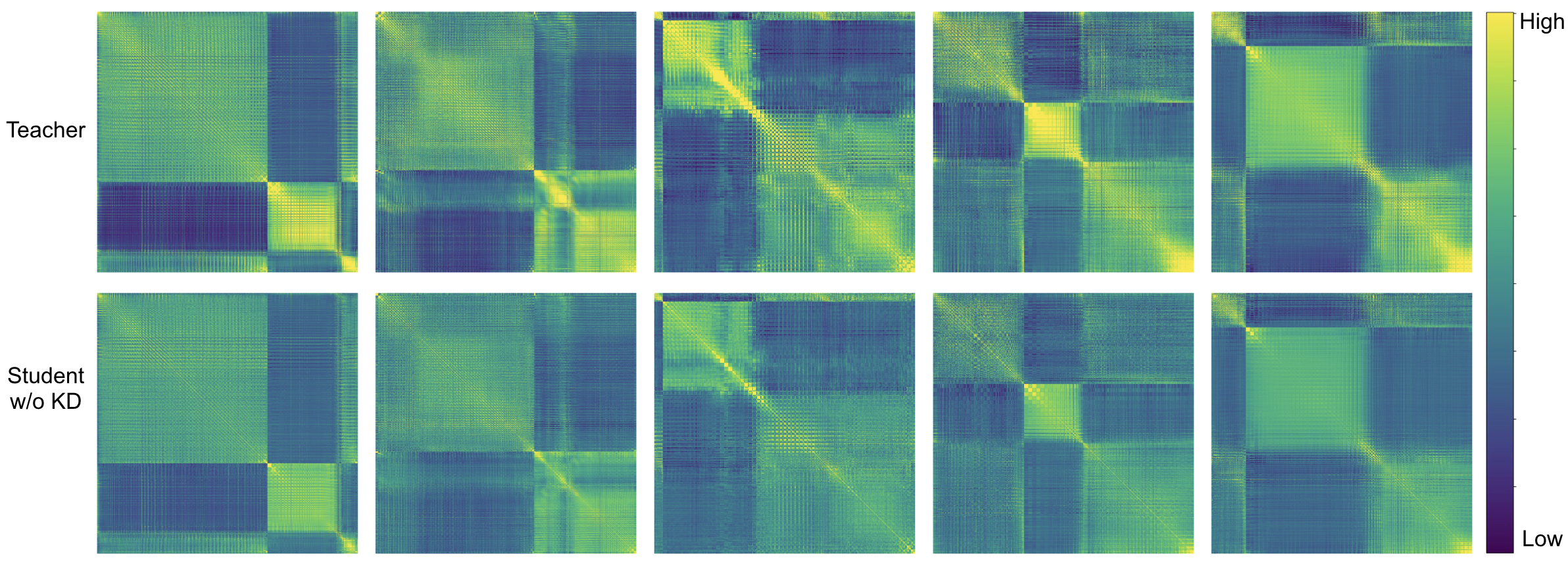}
\caption{To enhance the interpretability of this semantic similarity mapping, pixels are grouped and aligned together based on their semantic class. Brighter colors indicate a higher correlation. The teacher model exhibits similarity for semantic pixels within the same semantic class (diagonal block matrices) and dissimilarity across different semantic classes (off-diagonal block matrices). This matches with our hypothesis, where the teacher model displays clearer semantic relationship than the student model}

\label{activations}
\end{figure}
\textbf{Semantic Relation Activation Matrix.} Tung \& Mori \cite{tung2019similarity} demonstrated interestingly distinct activation patterns among image instances of different classes versus image instances of the same class. However, on the image-to-image translation tasks, less information is contained in instances' correlation as they are typically from the same class (e.g. horses, oranges). Our hypothesis is that similarity and dissimilarity might likewise present in the feature encoding of different semantic pixels, which is also more informative on the image-to-image translation tasks. A distillation loss can be introduced to penalize the difference between a teacher and a student's encoded similarity. We represent this activation matrix by the outer product of feature encoding $\mathcal{F}$, similar to \cite{zagoruyko2016paying,tung2019similarity}. Here, we define the feature encoding \textbf{$\mathcal{F}^{(i)}$} to be the output matrix of the \textit{i}-th image example at the last layer of encoder \textbf{$\hat E$}:

\begin{equation}
\begin{gathered}
\hat{ \mathcal{F}}_t^{(i)}= \hat{E}_t(x_i); \hat{\mathcal{F}}_s^{(i)} = \hat{E}_s(x_i),
\end{gathered}
\label{feature_encoding1}
\end{equation}
\begin{equation}
\begin{gathered}
\hat{\mathcal{F}}_t^{(i)}  \in \mathbb{R}^{1 \times C_t \times H' \times W'} \rightarrow  \mathcal{F}_t^{(i)} \in \mathbb{R}^{C_t \times (H' \cdot W')},\\
\hat{\mathcal{F}}_s^{(i)} \in \mathbb{R}^{1 \times C_s \times H' \times W'} \rightarrow  \mathcal{F}_s^{(i)} \in \mathbb{R}^{C_s \times (H' \cdot W')},
\end{gathered}
\label{feature_encoding2}
\end{equation}
where \textbf{$H'$} and \textbf{$W'$} indicate the feature encoding height and width while  \textbf{$C_t$}/\textbf{$C_s$} are number of channels respectively. We use a batch size of 1.
We then calculate semantic relation activation matrices $\mathcal{A} \in \mathbb{R}^{(H' \cdot W') \times (H' \cdot W')}$ as the outer product of $\mathcal{F}$, followed by a row-wise L2 normalization. 

\begin{equation}
\begin{gathered}
 \hat{\mathcal{A}_t} = \mathcal{F}_{t}^{(i)} \cdot {\mathcal{F}_{t}^{(i)}}^T;
 \hat{\mathcal{A}_s} = \mathcal{F}_{s}^{(i)} \cdot {\mathcal{F}_{s}^{(i)}}^T,\\
 \mathcal{A}_{t[k,:]} = \frac{\hat{\mathcal{A}_t} }{\sqrt{\sum_{j} {\hat{\mathcal{A}}^2_{t[k,j]}}}};
 \mathcal{A}_{s[k,:]} = \frac{\hat{\mathcal{A}_s} }{\sqrt{\sum_{j} {\hat{\mathcal{A}}^2_{s[k,j]}}}},
\end{gathered}
\label{attention_matrix}
\end{equation}

We show some evidence to support our intuition of semantic relation activation matrices in \textbf{Fig. \ref{activations}}. We sample 5 horse and zebra images from COCO dataset \cite{DBLP:conf/eccv/LinMBHPRDZ14} which provides ground truth segmentation masks, and generate all corresponding teacher's and student's activation matrices $\mathcal{A}$ by \textbf{equation \ref{attention_matrix}}. We group the values by pixels of the same semantic class to clearly show different activation patterns. The clear blockwise patterns in the teacher model indicate that pixels of the same semantic class are much more similar compared to pixels of different classes. On the other hand, this pattern is less observable in the student model learned without distillation. This empirical finding strongly supports our hypothesis that there exists certain relation patterns which can be explicitly transferred from a teacher network to a student network. Secondly, the activation matrix $\mathcal{A}$ is independent of the number of channels in feature $\mathcal{F}$, which avoids the difficulty of introducing a handcrafted feature loss to match $\mathcal{F}_t $ and $\mathcal{F}_s$ in different feature space. 
 
We define our semantic preserving distillation loss $\mathcal{L}_{SP}$ to be the L1 loss between two activation matrices:

\begin{equation}
\begin{gathered}
 \mathcal{L}_{SP} = {\mathbb{E}}_{\scriptscriptstyle{x \sim \mathcal{P}_{data}(x)}}\bigl[ \norm{\mathcal{A}_t - \mathcal{A}_s}_1\bigr],
\end{gathered}
\label{sp_loss}
\end{equation}

In preliminary experiments, we also tried L2 loss in enforcing the matching of two matrices but didn’t observe any significant improvement.
Our full objective is then,

\begin{equation}
\begin{gathered}
 \mathcal{L} = \mathcal{L}_{GAN_A}+\mathcal{L}_{GAN_B} + \gamma_1 \cdot \mathcal{L}_{SP_A} + \gamma_2 \cdot\mathcal{L}_{SP_B}\\
+ \lambda \Big(\alpha \mathcal{L}_{cyc}(G_s, F_s, X, Y) + (1 -  \alpha) \mathcal{L}_{cyc}(G_s, F_s, X_t, Y_t) \Big).
\end{gathered}
\label{full_loss}
\end{equation}
where $A$ and $B$ indicate the generators of each direction respectively. $\gamma_1$, $\gamma_2$ and $\alpha$ are hyper-parameters. $\lambda$ is the balancing coefficient.

\section{Experiments}

\subsection{Different image-to-image translation datasets}
\textbf{Setup.} To illustrate the effectiveness of our method on GAN compression, we qualitatively and quantitatively evaluated it on 5 benchmark image-to-image translation datasets including horse $\leftrightarrow$ zebra, summer $\leftrightarrow$ winter, apple $\leftrightarrow$ orange, tiger $\leftrightarrow$ leopard and Cityscapes label $\leftrightarrow$ photo.

We followed CycleGAN implementation and setup from the official PyTorch implementation\footnote{CycleGAN official PyTorch implementation: https://github.com/junyanz/pytorch-CycleGAN-and-pix2pix .} for a fair comparison. Specifically, the teacher generator stacks one 7x7 stride-1 convolutional layer, two 3x3 stride-2 convolutional layers, six or nine residual blocks, two 3x3 stride-2 transposed convolutional layers and one final 7x7 stride-1 convolutional layer sequentially. The student generator has the same architecture as the teacher generator but is narrower for each layer by a factor of 2 or 4 depending on the datasets trained on. Since all generators share the same structure in downsampling and upsampling parts, we use the number of residual blocks and the number of filters in the first convolutional layer to specify the generator architecture. This convention defines both depth and width of the model. Specifically, we used Resnet9, ngf64 and Resnet9, ngf16 as our major teacher student model pair for all datasets except horse $\leftrightarrow$ zebra dataset, where Resnet9, ngf32 is used for the student model.
As the Cityscapes dataset is inherently a paired dataset of the street view photo images and their corresponding semantic segmentation labels, we also conducted experiments in a Pix2Pix setting. The teacher and the student generators in our Pix2Pix experiments have a UNet structure \cite{isola2017image}. The discriminator network follows the PatchGAN discriminator \cite{isola2017image} structure. For all datasets, we trained and evaluated all models on images of resolution 256x256.

\begin{table}[b!]
  \centering
  \caption{The FID values for references/baselines (top) and variations of our methods (bottom). We conducted experiements on datasets horse-to-zebra (\emph{h} $\rightarrow$ \emph{z}, \emph{z} $\rightarrow$ \emph{h}), summer-to-winter (\emph{s} $\rightarrow$ \emph{w}, \emph{w} $\rightarrow$ \emph{s}), apple-to-orange (\emph{a} $\rightarrow$ \emph{o}, \emph{o} $\rightarrow$ \emph{a}), tiger-to-leopard (\emph{t} $\rightarrow$ \emph{l}, \emph{l} $\rightarrow$ \emph{t}). Lower is better. Both Co-evolutionary \cite{shu2019co} and ThiNet \cite{shu2019co} apply pruning while Co-evolutionary is specifically designed for compressing CycleGAN and ThiNet is a pruning method adapted from the classification task. For a fair comparison to Co-evolutionary and ThiNet, the models compared above have similar model size and computation requirement. (see \textbf{Table \ref{table:model_specs}}) In all cases except \emph{t} $\rightarrow$ \emph{l}, our method further improves over 50\% relatively compared to the baseline method (vanilla KD). FID trade off curve for varying ngf is in Supplementary}
  \begin{tabular}{ccccccccc}
    \toprule
    & \emph{h} $\rightarrow$ \emph{z} & \emph{z} $\rightarrow$ \emph{h} & \emph{s} $\rightarrow$ \emph{w} & \emph{w} $\rightarrow$ \emph{s} & \emph{a} $\rightarrow$ \emph{o} & \emph{o} $\rightarrow$ \emph{a} & \emph{t} $\rightarrow$ \emph{l} & \emph{l} $\rightarrow$ \emph{t} \\
    \midrule
    Teacher & 84.01 & 136.85 & 76.99 & 74.39 & 132.37 & 130.72 & 76.68 & 77.60 \\
     \midrule
    Student & 94.95 & 141.64 & \boldmath{$76.47$} & 74.90 & 132.99 & 137.10 & 93.98 & 89.37 \\
	ThiNet \cite{luo2017thinet} & 189.28 & 184.88 & 81.06 & 80.17 & - & - & - & - \\
    Co-evolutionary \cite{shu2019co}& 96.15 & 157.90 & 79.16 & 78.58 & - & - & - & - \\
    Vanilla KD & 106.10 & 144.52 & 80.10 & 79.33 & 127.21 & 135.82 & 82.04 & 87.29 \\
    \midrule
    Intermediate KD & 97.20 & 143.43 & 77.75 & \boldmath{$74.67$} & 126.90 & 133.16 & 86.82 & 92.99 \\
    + SP & 90.65 & 143.03 & 78.75 & 76.21 & 125.90 & 132.83 & 81.53 & 86.52 \\
    + 2 direction SP (\textbf{Ours}) & \boldmath{$86.31$} & \boldmath{$140.15$} & 76.59 & 75.69 & \boldmath{$121.17$} & \boldmath{$132.8$3} & \boldmath{$81.17$} & \boldmath{$80.75$} \\
    \bottomrule
  \end{tabular}

    \label{table:resnet9_fid}
\end{table}
\begin{table}[t!]
  \centering
	\caption{Computation and storage results for models on major experiments. T: teacher, S1, S2: student. Our models achieve superior performance in all tasks with a smaller/similar model size and computation compared to Co-evolutionary and ThiNet. We choose S1 on h $\leftrightarrow$ z and S2 on the rest of the datasets. The choice is made based on the gap between teacher and student baseline performance. Latency measurement plot for varying ngf is in Supplementary} 
  \begin{tabular}{ccccc}
    \toprule
    Model & Size (MB) & \#  Params & Memory (MB) & FLOPs \\
    \midrule
    ResNet 9blocks, ngf 64 (T) & 44 & 11.38M & 431.61 & 47.22G \\
    \midrule
    ThiNet \cite{luo2017thinet} & 11 (75\%$\downarrow$) & - & - & - \\
    Co-evolutionary \cite{shu2019co} h $\leftrightarrow$ z & 10 (77\%$\downarrow$) & - & - & 13.06G (72\%$\downarrow$) \\
    Co-evolutionary \cite{shu2019co} s $\leftrightarrow$ w & 7.6 (83\%$\downarrow$) & - & - & 10.99G (77\%$\downarrow$)  \\
    Co-evolutionary \cite{shu2019co} cityscapes & 12 (73\%$\downarrow$) & - & - & 16.45G (65\%$\downarrow$)  \\    
    ResNet 9blocks, ngf 32 (S1) & 11 (75\%$\downarrow$) & 2.85M (75\%$\downarrow$) & 216.95 (50\%$\downarrow$) & 12.14G (74\%$\downarrow$) \\
    ResNet 9blocks, ngf 16 (S2) & 2.8 (94\%$\downarrow$) & 0.72M (94\%$\downarrow$) & 109.62 (75\%$\downarrow$) & 3.20G (93\%$\downarrow$) \\
    \bottomrule
  \end{tabular}
  \label{table:model_specs}
\end{table}

\textbf{Quantitative Evaluation Metrics.} We adopt \textit{Fréchet Inception Distance} (FID) \cite{heusel2017gans} on horse $\leftrightarrow$ zebra, summer $\leftrightarrow$ winter, apple $\leftrightarrow$ orange and tiger $\leftrightarrow$ leopard datasets. FID calculates the Wasserstein-2 distance between feature maps extracted by Inception network from fake and real images. As a distance measure, a lower score is preferred for a higher correlation between synthetic and real images. 
On Cityscapes label $\rightarrow$ photo dataset \cite{cordts2016cityscapes}, we use FCN-score following the evaluation method used by Isola et al. \cite{isola2017image}. The method uses FCN-8s network, a pretrained semantic classifier, to score on synthetic photos with standard segmentation evaluation metrics from the Cityscapes benchmark including mean pixel accuracy, mean class accuracy and mean class Intersection over Union (IoU). 

\textbf{Quantitative Comparison.} In \textbf{Table \ref{table:resnet9_fid}}, we list our experiments conducted on 4 unpaired datasets trained on CycleGAN. We compare our results with two previous works \cite{shu2019co,luo2017thinet} on pruning and different settings of our design. As a reference to the compression ratio, we show a table of computed model size, the number of parameters, memory usage and the number of FLOPs in \textbf{Table \ref{table:model_specs}}.  

We explore variations of our methods on CycleGAN by conducting the following experiments: 1) We introduce an intermediate distillation loss on the fake image generated by the first generator in the cycle, computing an L1 norm difference between the teacher's generated image and the student's. We note this as \textbf{intermediate KD}. 2) We experiment with semantic relation preserving loss in two parts of the cycle. Semantic Preserving (\textbf{SP}) indicates that we only apply the semantic distillation loss on the first generator of the cycle (i.e. $\gamma_2=0$ in \textbf{equation (\ref{full_loss}})). \textbf{2 direction SP} denotes that we applied the semantic distillation loss on both generators in the cycle. '\textbf{+}' means it was added in addition to \textbf{Vanilla KD}.

Though all compared models reach a similar performance on the \emph{s} $\leftrightarrow$ \emph{w} dataset, our method accomplishes critically better performance than other methods on the rest of the datasets.  Adding our proposed distillation losses on both generators boosts the performance significantly from vanilla knowledge distillation, with the possibility to outperform the original teacher model on some tasks. We will further demonstrate visual evidence in later discussions. On the summer-to-winter task (\emph{s} $\leftrightarrow$ \emph{w}), however, we do not observe performance gain and we suspect the reason is that the baseline student model barely differs from the teacher model numerically. There is limited space and knowledge for improvement to take place.
Additionally, we run experiments on Cityscapes dataset and show FCN-score in \textbf{Table \ref{table:fcn_cyclegan}}. Interestingly, we notice a dramatic increase on FCN-score in applying the proposed method but only a similar or slightly better quality of image compared to the original model is observed (See Supplementary). Our proposed semantic preserving loss strongly reacts to this semantic segmentation dataset, by making pixels more recognizable in a semantic way.
\begin{table}[t!]
  \centering
  \caption{FCN-score for different models on the Cityscapes through CycleGAN training} 
  \begin{tabular}{cccc}
    \toprule
    & Mean Pixel Acc. & Mean Class Acc. & Mean Class IoU \\
    \midrule
    Teacher & 0.592 & 0.179 & 0.138 \\
    Student & 0.584 & 0.182 & 0.129 \\
    ThiNet \cite{luo2017thinet} & 0.218 & 0.089 & 0.054 \\
    Co-evolutionary \cite{shu2019co} & 0.542 & \boldmath{$0.212$} & 0.131 \\
    \textbf{Ours} & \boldmath{$0.704$} & 0.205 & \boldmath{$0.154$} \\
    \bottomrule
  \end{tabular}
  \label{table:fcn_cyclegan}
\end{table}

\begin{figure}[b!]
\includegraphics[width=\textwidth]{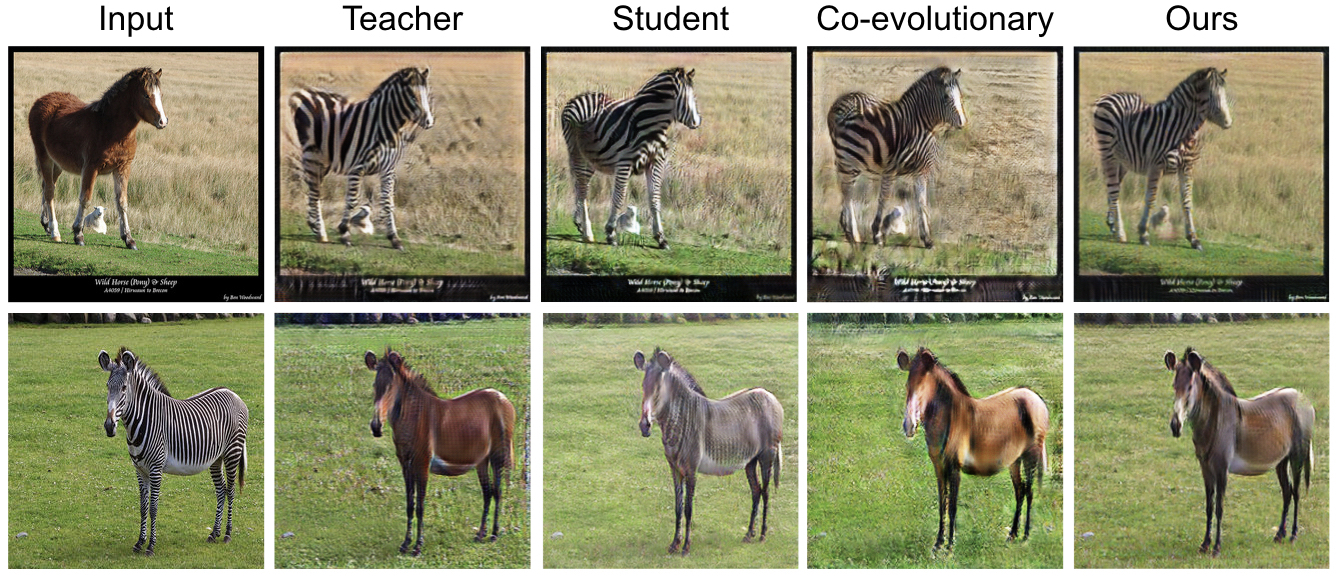}
\caption{Images generated by the teacher model, prior work \cite{shu2019co} and our proposed method on their selected examples. The top row displays input horse image and generated zebra; the bottom row displays input zebra image and generated horse images}
\label{huawei_comp}
\end{figure}

\begin{figure}[t!]
\centering
\includegraphics[width=0.8\textwidth]{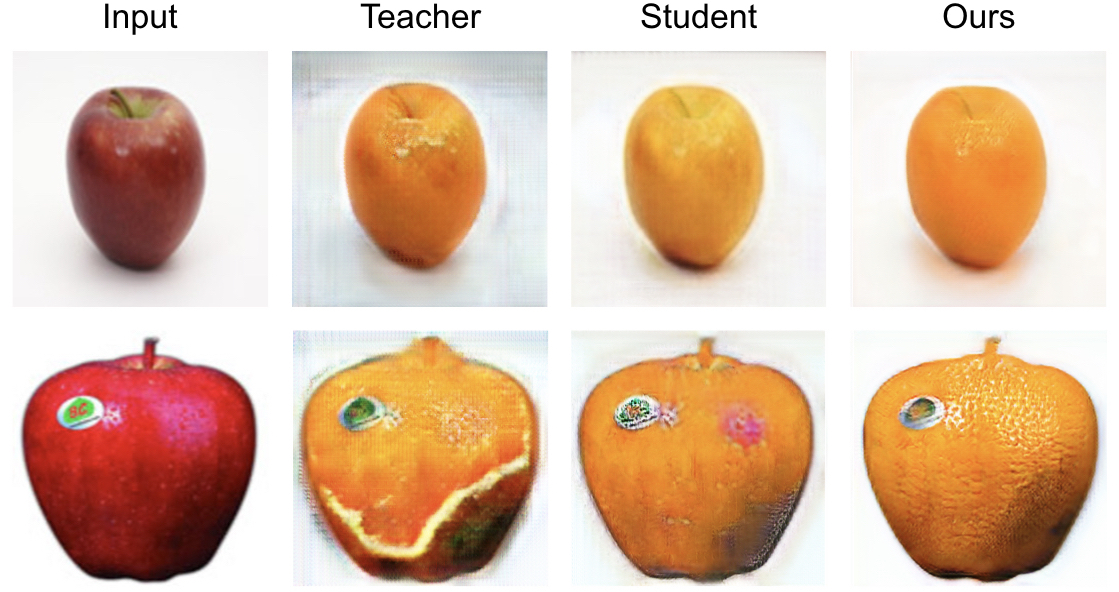}
\caption{Two examples of the apple-to-orange task. Clear and realistic texture is generated using our method, even outperforming the teacher}
\label{orange}
\end{figure}

\begin{figure}[t!]
\centering
\includegraphics[width=0.8\textwidth]{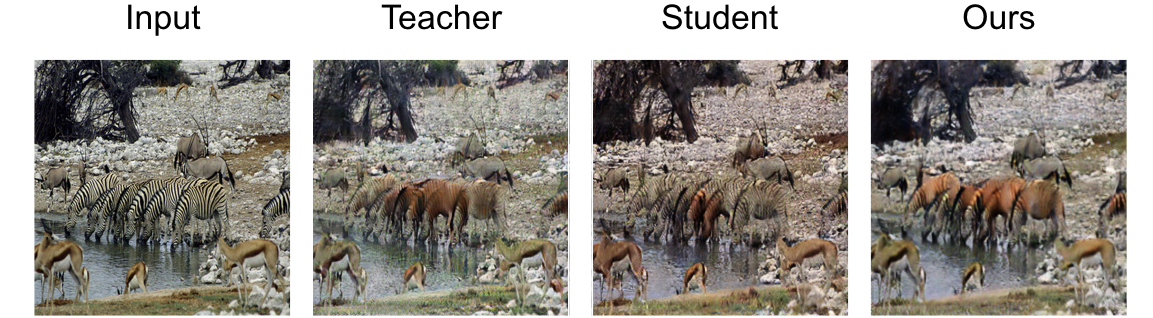}
\caption{SP method translated more zebras than the teacher and the student.}
\label{multi_horses}
\end{figure}

\begin{figure}[t!]
\centering
\includegraphics[width=1.0\textwidth]{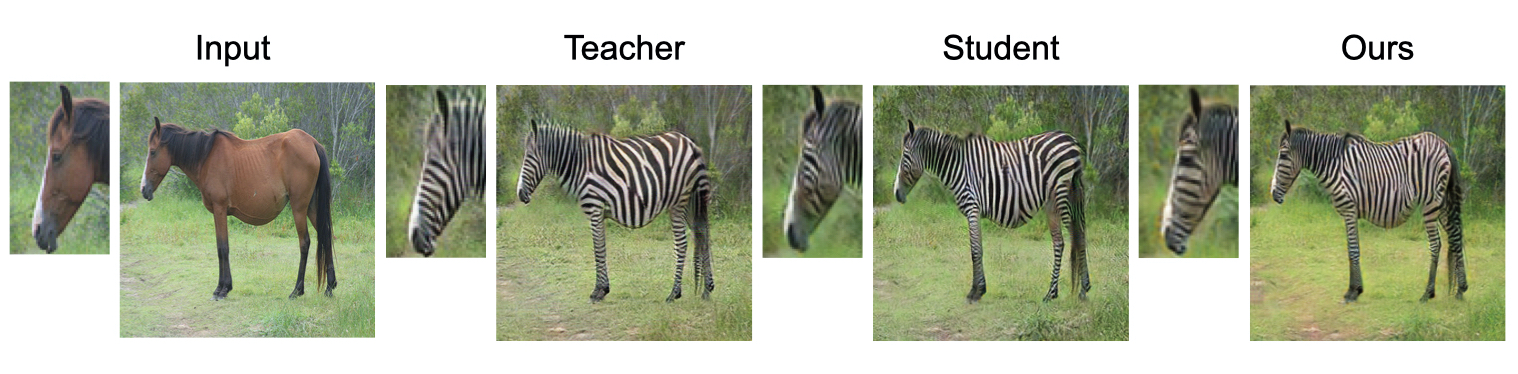}
\caption{An example on the horse-to-zebra task. A more detailed and realistic eye is preserved from horse to generated zebra image in our model}
\label{horse_eye_example}
\end{figure}

\textbf{Qualitative Results.} In this section, we present visual observations on the generated images from our models and reference models. To compare our results to \cite{shu2019co}, we also generated images using our models on their selected input images displayed in \textbf{Fig. \ref{huawei_comp}}. Evidently, our generated images contain a more realistic horse/zebra and reduce the artifacts to a minimum in the background.
In the earlier discussion, we mention the potential of the student model to outperform the teacher by adding our proposed semantic preserving loss, with the numerical evidence in \textbf{Table \ref{table:resnet9_fid}}. The extra guidance signal from the teacher’s pairwise semantic preserving activations not only encourages the student to learn more intra-pixel relationships within a specific image but also semantic understanding of the entire training population. Furthermore, this method accelerates the learning of discriminators towards catching more details in the early stage. Incorporating both effects empowers the student model to even outperform the teacher model in certain cases. In \textbf{Fig. \ref{orange}}, we show 2 significant examples where our proposed method achieves exceptionally better results.

\begin{figure}[t!]
\centering
\includegraphics[width=0.99\textwidth]{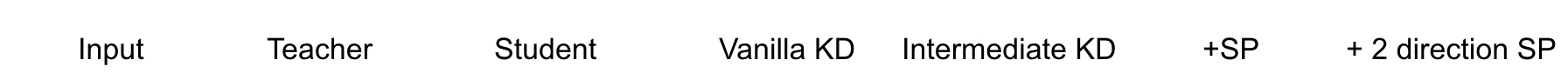}
\includegraphics[width=0.99\textwidth]{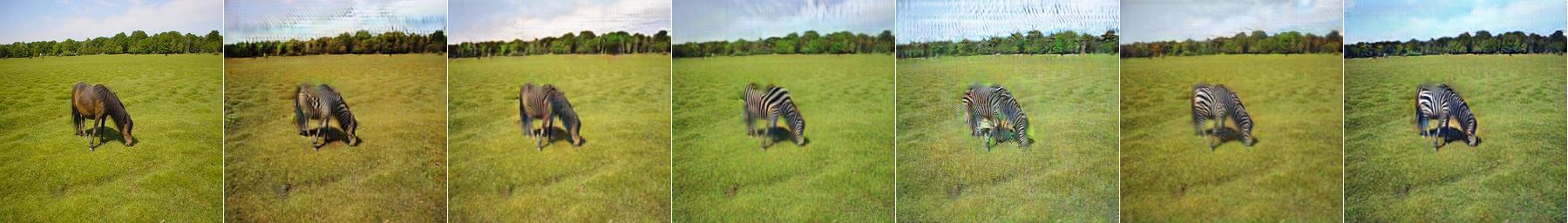}
\includegraphics[width=0.99\textwidth]{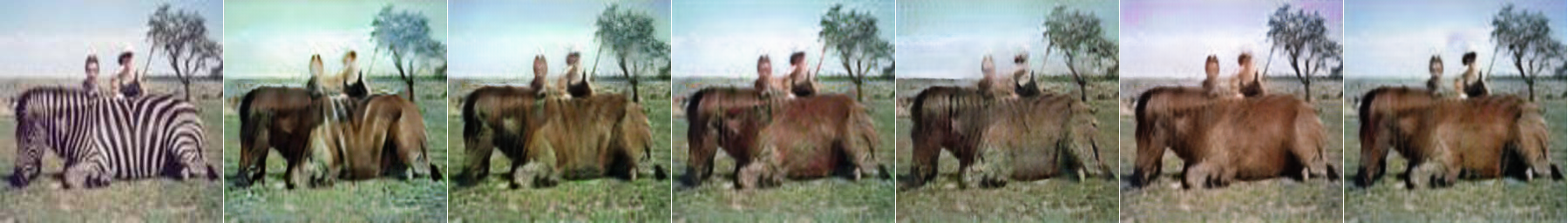}
\includegraphics[width=0.99\textwidth]{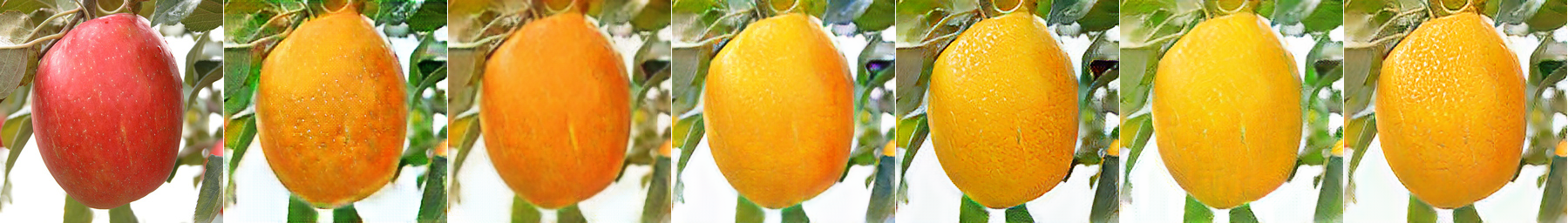}
\includegraphics[width=0.99\textwidth]{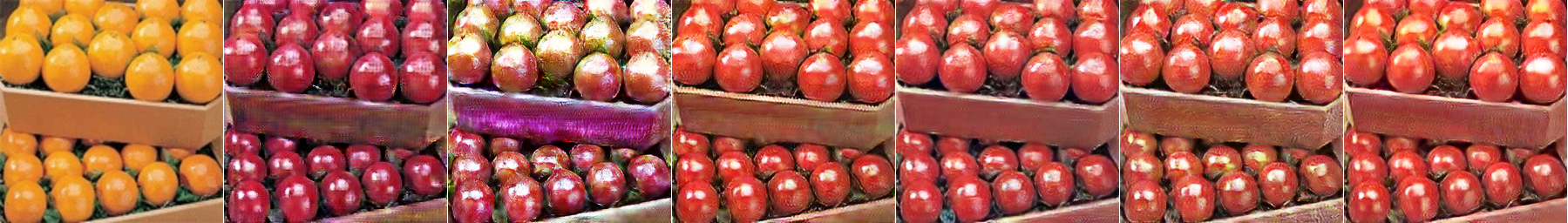}
\includegraphics[width=0.99\textwidth]{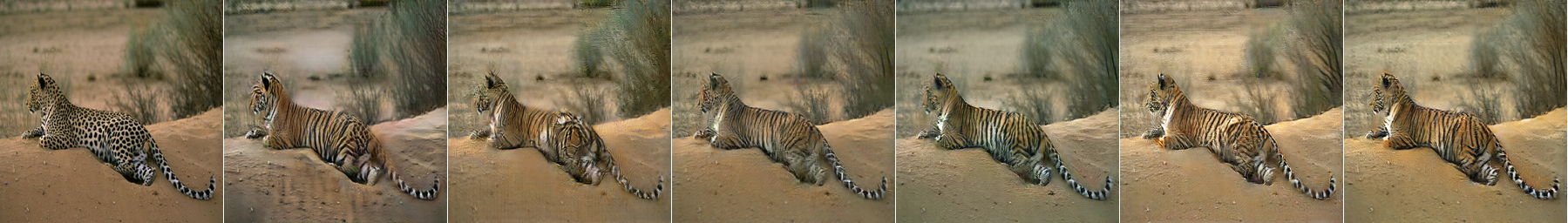}
\includegraphics[width=0.99\textwidth]{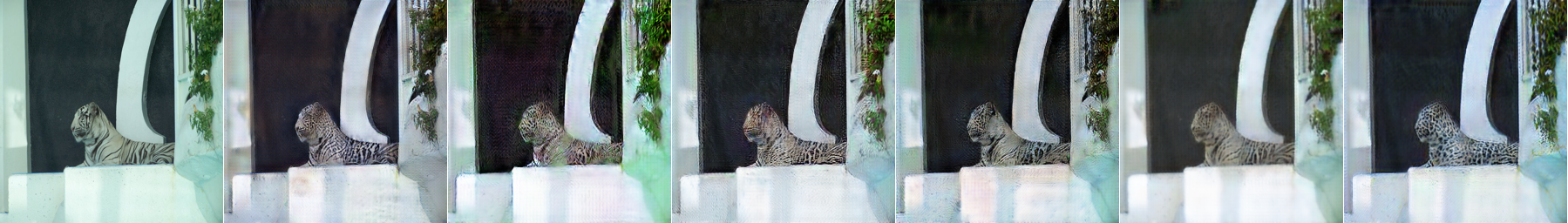}
\includegraphics[width=0.99\textwidth]{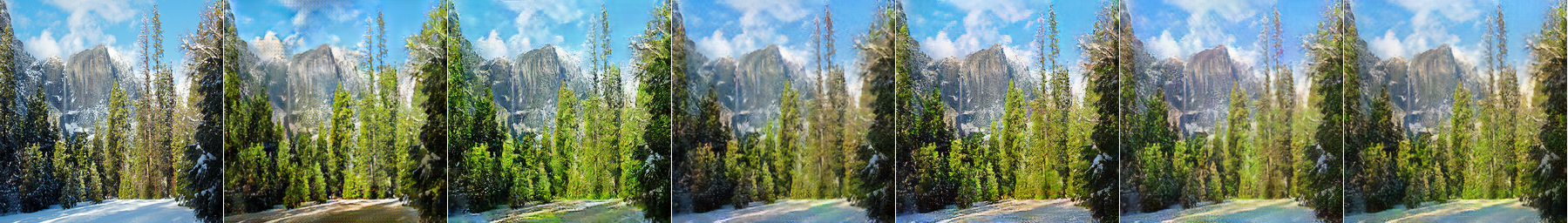}
\includegraphics[width=0.99\textwidth]{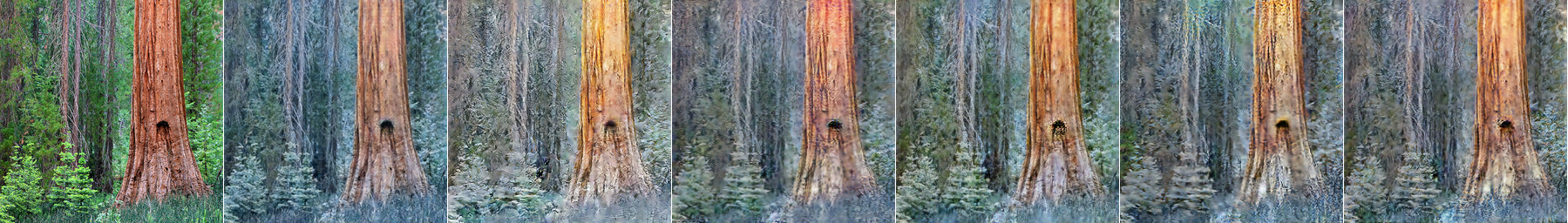}
\caption{Ablation study: examples from multiple datasets comparing results in baseline models and variation of our methods}
\label{multi_datasets}
\end{figure}

An intriguing example failure (\textbf{Fig. \ref{multi_horses}}) case shows that CycleGAN fails to transfer multiple objects while our method succeeds. We also observe significantly better details and textures preserving in different tasks. In \textbf{Fig. \ref{horse_eye_example}}, a more detailed and realistic eye is preserved. We also provide examples from different datasets in \textbf{Fig. \ref{multi_datasets}} in an ablation study manner.

\subsection{Different architectures}
We also demonstrate that our method is extensible to other types of network structure. On the horse-to-zebra task, we test Resnet6 generator for both the teacher and the student models. The FID evaluation is shown in \textbf{Table} \ref{table:fid_resnet6}. Our method's performance gain the most over others but less significant as in the Resnet9 case. We conjecture that the teacher model of 6 blocks contains less semantic relationships, which limits the amount of knowledge to be transferred. We also conducted additional experiments using UNet \cite{isola2017image} on the Cityscapes dataset translate from semantic mask to street view photos. As UNet's encoder downsamples the input to 1x1 resolution at the bottleneck layer, the desired spatial semantic information is lacking at the bottleneck layer. Therefore, we distill the semantic relation activation matrix at layer 2 and layer 3. We show FCN-score results in \textbf{Table \ref{table:fcn_pix2pix}}.  We find the highest mean pixel accuracy with distilling at layer 3 but similar 
mIoU at both layers. Compression ratio and visual results can be found in Supplementary.

\begin{table}[t!]
  \caption{FID value for Resnet6 generators on Horse $\leftrightarrow$ Zebra dataset} 
  \centering
  \begin{tabular}{ccccccc}
    \toprule
    & Teacher & Student & Intermediate KD & Vanilla KD & + SP & + 2 direction SP \\
    \midrule
    \emph{h} $\rightarrow$ \emph{z} & 88.27 & 109.93 & 107.04 & 105.49 & 108.71 & \textbf{105.51} \\
    \emph{z} $\rightarrow$ \emph{h} & 143.08 & 144.01 & 142.63 & 146.26 & \textbf{141.50} & 141.90 \\
    \bottomrule
  \end{tabular}
  \label{table:fid_resnet6}
\end{table}

\begin{table}[t!]
  \caption{FCN-score on Cityscapes. Feature encoding extracted from Unet256 has a spatial resolution of 64x64 and 32x32 at layer 2 and layer 3 respectively}
  \centering
  \begin{tabular}{cccc}
    \toprule
    & Mean Pixel Acc. & Mean Class Acc. & Mean Class IoU \\
    \midrule
    Teacher & 0.757 & 0.235 & 0.191 \\
    Student & 0.710 & 0.219 & 0.169 \\
    Vanilla KD & 0.742 & 0.224 & 0.182 \\
    \midrule
    + SP layer 2 & 0.743 & \boldmath{$0.230$} & \boldmath{$0.183$} \\
    \textbf{+ SP layer 3} & \boldmath{$0.770$} & 0.229 & \boldmath{$0.183$} \\
    \bottomrule
  \end{tabular}
  \label{table:fcn_pix2pix}
\end{table}

\section{Conclusions}
We approach model compression of GANs via a novel proposed method extended on traditional knowledge distillation. Our strategy, which transfers semantic relation knowledge from a teacher model to a selected student model, shows strong potential in generating images with better details and texture after explicitly leaning the semantic relationships while using knowledge distillation to significantly reduce the model size and computation requirement. Our experiments conducted on 5 different datasets and 3 different architectures have demonstrated quantitatively and qualitatively that our proposed method helps bring a previously incompetent student network to the level of its teacher, with the capability to generate images at a significantly higher level of quality. 

\textbf{Acknowledgement} Authors thank Brendan Duke, Soheil Seyfaie,  Zhi Yu, Yuze Zhang for their comments and suggestions.

\clearpage
%
%
\bibliographystyle{splncs04}
\bibliography{egbib}
\clearpage
\section{Supplementary}

In this supplementary material, we provide more details on the network latency measurement and FID trade off curve. We also include objective equation for Pix2Pix setup and its experimental results on Cityscapes dataset.
\subsection{Latency Measurement for Different Configurations}

\begin{figure}[!htb]
 \centering
 \begin{tikzpicture}
   \begin{axis}[
     width=12cm,
     height=8cm,  
     grid=both,
     xmin=0,
     xmax=140,
     ymin=0,
     ymax=10,
     xlabel={ngf},
     ylabel={Second (s)}
   ]
     \addplot[mark=*,black] plot coordinates {
       (8,0.21)
       (16,0.43)
       (32,1.01)
       (64,2.69)
       (128,8.21)
     };
   \end{axis}
 \end{tikzpicture}
 \caption{Latency measurement on a single CPU core of Intel(R)
Xeon(R) E5-2686 with different ngfs for Resnet based generators on CycleGAN Experiment}
\end{figure}
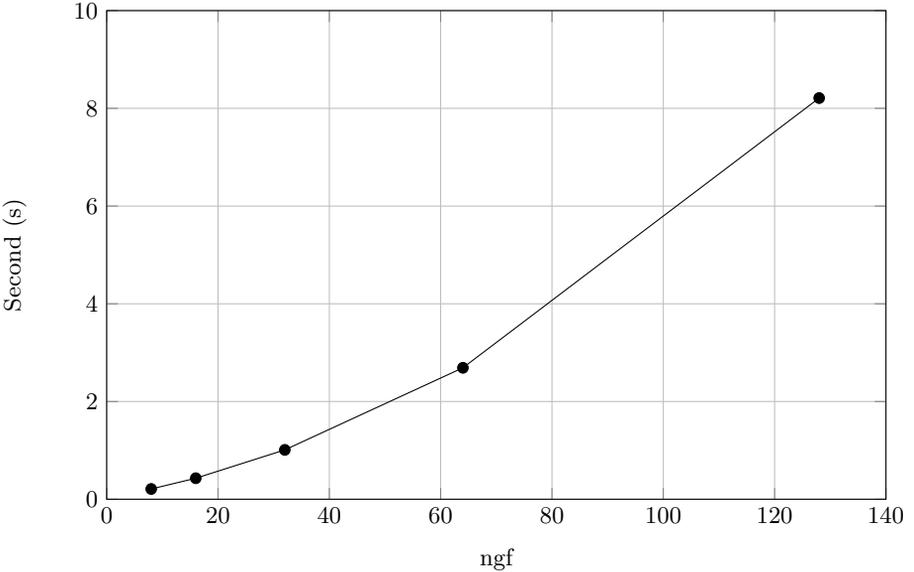

\subsection{FID Trade off Curve for Different Configurations for CycleGAN Experiment}

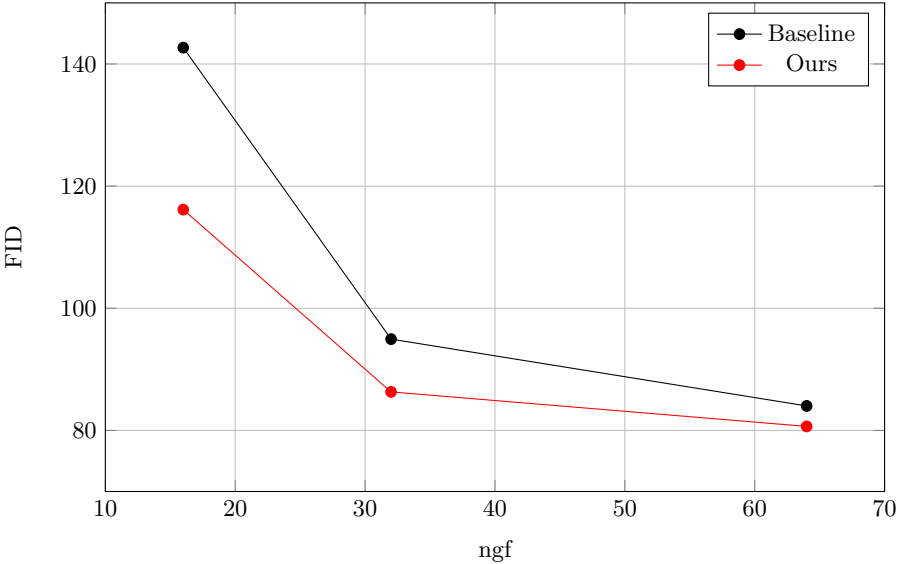
\begin{figure}[!htb]
 \centering
 \begin{tikzpicture}
   \begin{axis}[
     width=12cm,
     height=8cm,  
     grid=both,
     xmin=10,
     xmax=70,
     ymin=70,
     ymax=150,
     xlabel={ngf},
     ylabel={FID}
   ]
     \addplot[mark=*,black] plot coordinates {
       (16,142.67)
       (32,94.95)
       (64,84.01)
     };
     \addlegendentry{Baseline}
     \addplot[mark=*,red] plot coordinates {
       (16,116.15)
       (32,86.31)
       (64,80.66)
     };
     \addlegendentry{Ours}
   \end{axis}
 \end{tikzpicture}
 \caption{FID trade off curve on \emph{h} $\rightarrow$ \emph{z} for CycleGAN Experiment}
\end{figure}

\subsection{Knowledge Distillation Objective Function for Pix2Pix Experiment}
\textbf{Vanilla Knowledge Distillation.} Different from CycleGAN setting, which involves two generators and cycle consistency loss, Pix2Pix only does one direction translation trained with paired data in a supervised way. Analogy to how vanilla knowledge distillation applied on classification task, the objective function has the following form:
\begin{multline}
\mathcal{L}(G_s, D) = \lambda \bigl( \alpha \cdot \mathcal{L}_{L1}(G_s, X, Y) + (1 -  \alpha) \cdot \mathcal{L}_{L1}(G_s, X, Y_t)\bigr) \\
+ \mathcal{L}_{GAN}(G_s, D, X, Y),
\end{multline}
where $\mathcal{L}_{L1}$ is an L1 norm loss between the ground-truth labels and the generated images. $\lambda$ is the balancing coefficient for $\mathcal{L}_{L1}$. $\alpha$ is the hyper-parameter to weigh between the true label and the teacher's label.

\textbf{Semantic Preserving Knowledge Distillation.} Built on vanilla knowledge distillation objective, semantic preserving knowledge distillation loss is directly added to the above objective function:

\begin{equation}
\begin{gathered}
 \mathcal{L} = \mathcal{L}_{GAN} + \gamma \cdot \mathcal{L}_{SP} + \lambda \bigl( \alpha \cdot \mathcal{L}_{L1}(G_s, X, Y) + (1 -  \alpha) \cdot \mathcal{L}_{L1}(G_s, X, Y_t)\bigr).
\end{gathered}
\end{equation}

\subsection{Model Size and Computation Results for Pix2Pix Experiment}
The teacher and the student models used in Pix2Pix experiments with computation and storage statistics are shown in \textbf{Table \ref{table:unet_specs}}. 
\begin{table}[]
  \centering
	\caption{Computation and storage results for models on Pix2Pix experiments. The choice is made based on the gap between the teacher and the student baseline performance} 
  \begin{tabular}{ccccc}
    \toprule
    Model & Size (MB) & \#  Params & Memory (MB) & FLOPs \\
    \midrule
    UNet256, ngf 64 (T) & 208 & 54.41M & 51.16 & 2.03G \\
    UNet256, ngf 16 (S) & 11(95\%$\downarrow$) & 3.40M(94\%$\downarrow$) & 13.91(73\%$\downarrow$) & 0.14G(93\%$\downarrow$) \\
    \bottomrule
  \end{tabular}
  \label{table:unet_specs}
\end{table}

\subsection{Qualitative Results on Cityscapes}
\begin{figure}[t!]
\includegraphics[width=\textwidth]{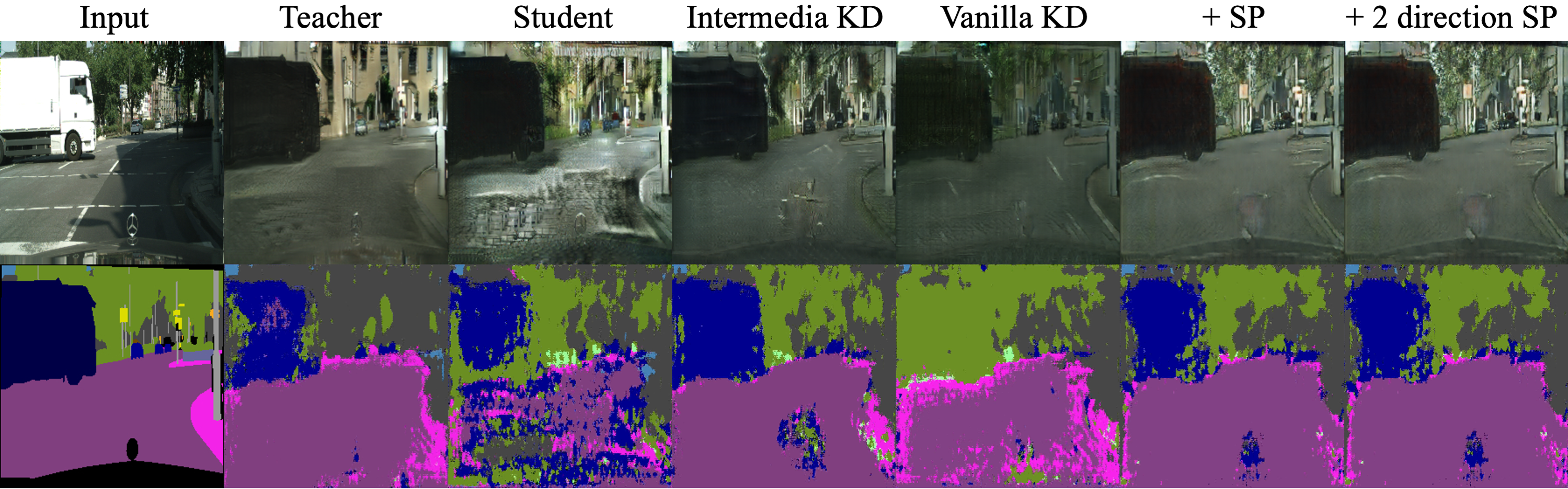}
\caption{Ablation study: generated street view images with FCN-8s segmented masks through CycleGAN training on the Cityscapes dataset. 
The image generated by our method (last column) significantly reduces artifacts compared to the student's generated image. Although the teacher generates a more realistic image, we observe that our model preserves pixels' semantic class with respect to the input mask. For example, in the top right corner, the teacher's generated image only includes buildings in the designated region of trees}
\label{cityscapes_cyclegan}

\end{figure}

\begin{figure}[t!]
\includegraphics[width=\textwidth]{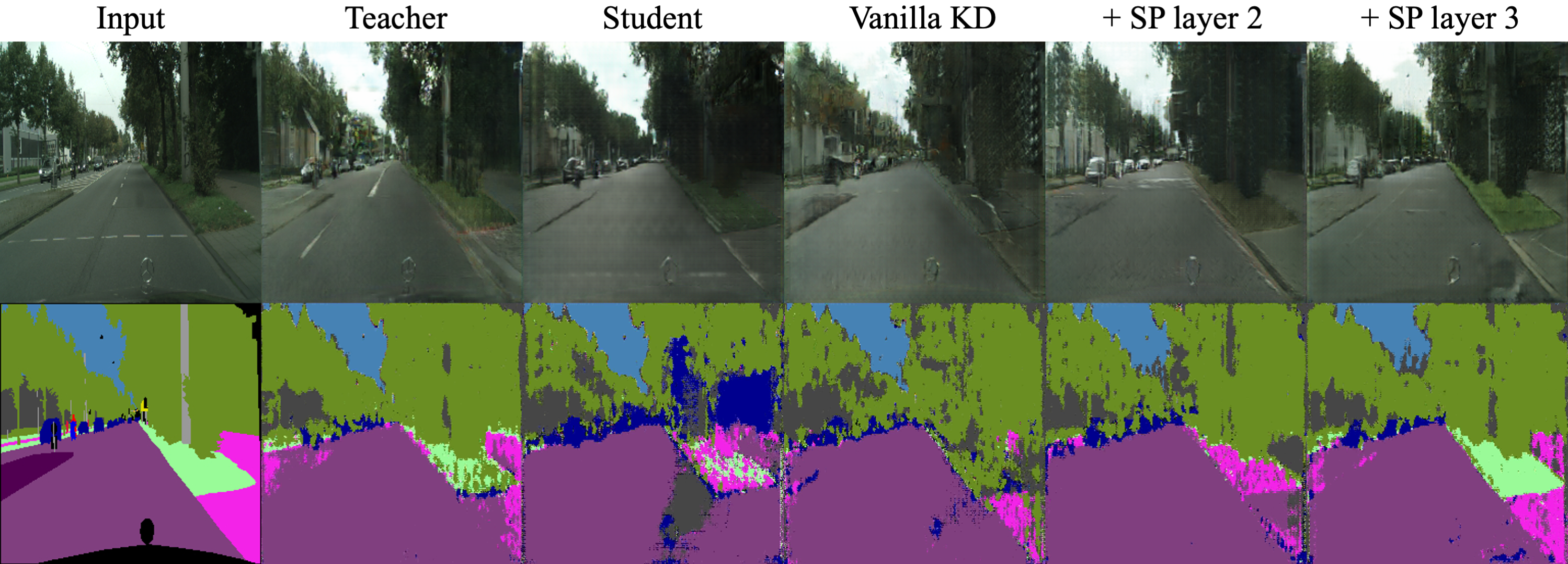}
\caption{Ablation study: generated street view images with FCN-8s segmented masks through Pix2Pix training on the Cityscapes dataset. Among all generated masks, our model (last column) shows the most distinct segmentation mask with clear boundaries of each semantic class. For instance, on the right of the segmented masks, we observe a significant improvement at the boundary of the green belt and the side walk.}
\label{cityscapes_pix2pix}

\end{figure}

On the Cityscapes dataset, we conducted both paired and unpaired image translation experiments via Pix2Pix and CycleGAN training, respectively. The synthetic street view images translated from their semantic masks along with FCN-8s generated instance segmentation masks are displayed in \textbf{Fig \ref{cityscapes_cyclegan}} and \textbf{Fig \ref{cityscapes_pix2pix}}.

\subsection{Experiment Details}
All models are trained on 256x256 input images with a batch size of 1 and optimized using Adam. The other settings for GAN training is the same as CycleGAN and Pix2Pix. 

horse $\leftrightarrow$ zebra, summer $\leftrightarrow$ winter and apple $\leftrightarrow$ orange datasets are downloaded using CycleGAN provided script. horse $\leftrightarrow$ zebra with segmentation mask sample image, which is used to draw semantic similarity matrices is downloaded from COCO \cite{DBLP:conf/eccv/LinMBHPRDZ14}. tiger $\leftrightarrow$ leopard dataset is obtained from ImageNet \cite{deng2009imagenet} using keyword \textit{tiger} and \textit{leopard}. The Cityscapes dataset is download from the official website \footnote{Official Cityscapes website: https://www.cityscapes-dataset.com/}. 

Implementation of FID score is adapted from a PyTorch port version of its official implementation \footnote{PyTorch port version of its official implementation: https://github.com/mseitzer/pytorch-fid}. Calculation of FCN-score is provided in Pix2Pix official Torch implementation \footnote{Pix2Pix Torch official implementation: https://github.com/phillipi/pix2pix}.

In the vanilla knowledge distillation training, we set $\lambda=10$ and $\alpha=0.05$ for all experiments. $\gamma$ ($\gamma_1 = \gamma_2$) is set to 0.9 in horse $\leftrightarrow$ zebra, 0.5 in summer $\leftrightarrow$ winter, 0.8 in apple $\leftrightarrow$ orange, 0.2 in tiger $\leftrightarrow$ leopard and 0.2 in Cityscapes for the unpaired translation experiments. In the paired translation experiments, $\gamma$ is set to 1 and $\lambda$ is set to 100.

\end{document}